# How the Symbol Grounding of Living Organisms Can Be Realized in Artificial Agents

**J. H. van Hateren**                                                    J.H.VAN.HATEREN@RUG.NL
*Johann Bernoulli Institute for Mathematics and Computer Science*
*University of Groningen, P.O. Box 407, 9700 AK Groningen, The Netherlands*

### Abstract

A system with artificial intelligence usually relies on symbol manipulation, at least partly and implicitly. However, the interpretation of the symbols – what they represent and what they are about – is ultimately left to humans, as designers and users of the system. How symbols can acquire meaning for the system itself, independent of external interpretation, is an unsolved problem. Some grounding of symbols can be obtained by embodiment, that is, by causally connecting symbols (or sub-symbolic variables) to the physical environment, such as in a robot with sensors and effectors. However, a causal connection as such does not produce representation and aboutness of the kind that symbols have for humans. Here I present a theory that explains how humans and other living organisms have acquired the capability to have symbols and sub-symbolic variables that represent, refer to, and are about something else. The theory shows how reference can be to physical objects, but also to abstract objects, and even how it can be misguided (errors in reference) or be about non-existing objects. I subsequently abstract the primary components of the theory from their biological context, and discuss how and under what conditions the theory could be implemented in artificial agents. A major component of the theory is the strong nonlinearity associated with (potentially unlimited) self-reproduction. The latter is likely not acceptable in artificial systems. It remains unclear if goals other than those inherently serving self-reproduction can have aboutness and if such goals could be stabilized.

## 1. Introduction

Much of artificial intelligence relies on manipulating symbols. Even a system that relies mostly on connectionist processing of sub-symbolic variables (e.g., Brooks, 1991) must ultimately be interpreted symbolically, such as when a human judges its performance in terms of whether it relates inputs to outputs in a meaningful way. Transforming meaningful inputs into meaningful outputs is a symbolic transformation. Newell and Simon (1976) conjectured that information processing by a physical symbol system is all that is needed for producing general intelligent action of the kind humans produce. Physical symbols, such as present in the physical patterns within a computing machine, can be regarded as forming a formal system. Searle (1980) questioned the sufficiency of formal systems for producing some of the major characteristics that symbols, and combinations of symbols, have for humans. Symbols, as used for example in human language, typically refer to something else that does not need to resemble the symbol. For example, the word "tree" can refer to an actual tree that has no resemblance to the word. Symbols therefore refer in an arbitrary way, only agreed upon by convention. In addition, human symbols are typically vague and ambiguous, with their meaning shifting depending on context. The word "tree" can mean quite different things to a forester, a genealogist, or a computer scientist. Even within a particular field, symbols get many different interpretations depending on context. But the most puzzling aspect of symbols is their capability to be about something else, even if this something else is not concrete like a specific tree, but quite abstract like a tree in general or very abstract like the concept "danger". The aboutness (the being about and referring to) of symbols as used by humans, and the apparent lack of aboutness of symbols in artificial systems, was Searle's main argument to question the sufficiency of physical symbol systems for obtaining general, human-level intelligence.

Harnad (1990) discussed the arbitrariness of symbols in a purely formal system as the "symbol grounding problem", the problem that in a formal system it is not clear how symbols can unequivocally connect to physical reality. The proposed solution was to use a connectionist system as intermediate between the formal system and sensory data. A material system with sensors and possibly effectors, as in a robot, can produce a causal connection between external reality and internal symbolic processing, by embodying the symbols. In a way, this was already discussed as the "robot reply" by Searle (1980, p. 420), and deemed not sufficient for solving the problem of aboutness (also known as "intentionality" in philosophical jargon). "Symbol grounding" as defined by Harnad (1990) is focused on physical reference, which does not specifically aim to solve the general problem of aboutness (Rodríguez et al., 2012). In this article I use "symbol grounding" in a wider sense (following Sun, 2000), thus including the grounding of reference (aboutness) in the more general sense. Symbol grounding for artificial systems then becomes the problem of how to produce reference for and meaning within an artificial agent without meaning being parasitic on human interpretation. Although many solutions have been proposed, often by combining symbolic with connectionist processing, none appear to really solve the problem of aboutness (for reviews see Ziemke, 1999; Taddeo & Floridi, 2005; Rodriguez et al., 2012; Coradeschi et al., 2013; Bielecka, 2015).

Recent theoretical and computational work has produced a plausible explanation of how aboutness and symbols have evolved in living organisms in general and humans in particular (van Hateren, 2014, 2015). Below I will start with reviewing this theory, and explain the mechanisms involved. I will then abstract these mechanisms from their biological context, and formulate them in a form that is independent of a specific material implementation. I will subsequently discuss the prospects for copying such mechanisms into artificial agents, and identify the primary problems that would need to be solved for accomplishing that.

## 2. Biological Symbol Grounding

Understanding how aboutness has likely arisen in organic evolution requires appreciating three key notions. The first is that the basic Darwinian theory of differential reproduction (organisms equipped to reproduce more effectively than others are likely to become dominant, that is, they appear to be naturally selected) has an extension that depends on the organism establishing an internal model of its capability to reproduce. The second notion is that the specific, stochastic role of this internal model produces a connection between internal and external variables that is indirect; it represents rather than causally connects, thus producing a primordial form of aboutness. The third notion is that communication between organisms can reduce the ambiguity that results from combining such internal variables into abstract symbols. The three sections below explain these notions.

## 2.1 Extending Basic Darwinian Evolution

The basic Darwinian theory of evolution by natural selection consists of a reproductive loop (loop $R$ in Figure 1) where an organism (called agent below) produces offspring that incorporates slight structural changes. The rate of reproducing is given by a quantity called fitness ($f_{true}$ in the figure). Fitness is taken here as an expected rate of reproduction (i.e., number of offspring per unit of time), with the actual number of offspring the stochastic (random) realization of the rate as integrated over the lifetime of the agent. In evolutionary biology, fitness is often measured, empirically and retrospectively, as a simple parameter estimated by counting offspring. However, for the purpose of modeling and prediction, fitness should be seen as the result of a complex dynamical process, also denoted by the term fitness (then more akin to a function, which has a value, form, and argument(s)). This fitness process has the time-varying properties of the environment and the agent (including its memory) as inputs and the expected reproduction rate as output. The output will vary continually and instantly, for example temporarily dropping at times when resources are scarce, recovering when resources become abundant, and finally going to zero when the agent dies. The structure (form) of $f_{\text{true}}$ can be extremely complex for complex agents in a natural environment. When $f_{\text{true}}$ or its internal estimate are mentioned below, this refers either to their value or to their structure, which either will be clear from the context or will be specified explicitly.

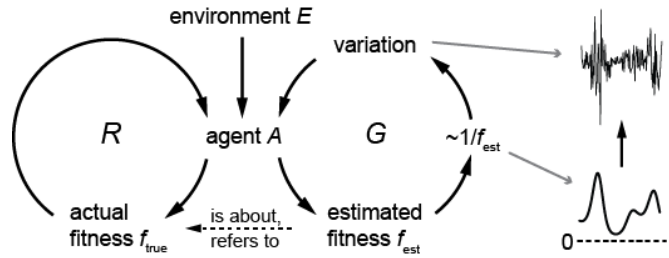

Figure 1: Basic Darwinian theory and extension; agents reproduce in a reproductive loop ($R$) depending on a reproduction rate $f_{\text{true}}$, while an agent-made estimate $f_{\text{est}}$ of $f_{\text{true}}$ stochastically drives structural, behavioral changes (hereditary and learned) in an agent's goal-directed loop ($G$); $f_{\text{est}}$ is about $f_{\text{true}}$

Agents typically occupy an environment that continually and unpredictably changes and they compete with one another because of limited resources, such as limited materials, limited energy, and limited space. Environmental change and competition tend to drive fitness down, and only by changing, and thereby improving their fitness, agents and lineages can survive. Such changes can take place not only over evolutionary time (across generations, such as in the form of mutations), but also within the lifetime of a single agent through learning, resulting in behavioral change. Although the latter changes usually are not transferred to offspring (unless there is cultural transfer), learning still influences fitness as defined here as an instantly varying rate of reproduction. It is important to stress that the power of the Darwinian process originates from the fact that reproduction is a multiplicative process: when fitness is above the replacement level, growth in numbers will be very fast (exponential). On the other hand, when fitness remains below the replacement level for a particular line of descending agents, that lineage will eventually become extinct.

In addition to the basic Darwinian process, directly depending on $f_{\text{true}}$, there is a second process that can promote evolutionary success by utilizing fitness indirectly (van Hateren, 2015). This is illustrated by the $G$ loop in Figure 1. The basic idea is that some of the structural changes in agents (either across generations or within their lifetime) are controlled by an internal estimate of their own fitness produced within the agents themselves. Such an estimate, called $f_{\text{est}}$ in Figure 1, subsequently drives the variance of structural change (called "variation" in the figure) in an inverse way depending on the value of $f_{\text{est}}$ (symbolized by ~$1/f_{\text{est}}$ in the figure; the precise form is subject to evolutionary optimization). The traces to the right illustrate how this works: a non-negative variable (lower trace) modulates the variance of a stochastic process (upper trace). This way of causing change is therefore midway between regular deterministic causation (the usual way things are modeled in science) and regular stochastic causation (e.g., letting a random generator drive downstream effects). The key point here is that the noise is not added to a deterministic variable, but is multiplied by it.

As noted above, fitness can best be seen as a process (in the case of $f_{est}$ a physiological process within the agent), with as output a rate of reproduction (in the case of $f_{est}$ an estimated rate of reproduction). However, the value of $f_{est}$ is taken to be present not explicitly (as a specific physiological variable) but only implicitly, in a distributed way within the agent. Similarly, producing structural change in the agent is also seen as a distributed process. When there are many different structures (behaviors) that could be changed, then the partial fitness effects of the inputs to $f_{est}$ and the outputs to structural change would need to be taken into account such that the changes are weighted properly. The required system would be very hard to design from first principles, but should be readily amenable to evolutionary optimization.

The theory has been shown to work in evolutionary simulations (van Hateren, 2015), for structural changes at evolutionary as well as agential timescales. The reason it works can be understood intuitively as follows. When $f_{true}$ is high, $f_{est}$ is likely to be high as well because it is assumed to be a reasonable estimate of $f_{true}$. When fitness is high, things are going well for the agent, and there is no reason to change much – just a little variation would be beneficial to facilitate future adaptation to environmental change. Thus when fitness is high, variation should be low. But when fitness is low, the agent is in trouble, and not changing (thereby likely maintaining low fitness for the agent and its descendants) is likely to lead to death and extinction, at least on average. Then it is a better strategy to increase variation in the hope of finding a variant with higher fitness ("desperate times call for desperate measures"). Because the variation is random, there is a considerable chance that the change will be for the worse, subsequently increasing variation even more (the $G$ loop acts continually, and a change for the worse will decrease $f_{est}$ and thereby increase variation). However, when a variant with higher fitness is encountered by chance, variation will be strongly reduced (because of the $\sim 1/f_{est}$). Then the agent will remain close to this beneficial state, drifting away only slowly, at least until environmental change reduces the fitness and things start over again. The $G$ loop is therefore an adaptive optimization process. The reason it works is that the low probability of success (when changing randomly) is compensated, on average, by the prospect of exponential growth when fitness becomes high enough.

The mechanism of the $G$ loop uses only random change and thus only concerns structural change where the effects on fitness cannot be foreseen. Yet, there are many cases where the effects of structural change can be foreseen, at least partly. Foresight is possible, for example, when knowledge about similar situations in the past is available in genetic memory (as established through previous natural selection) or in behavioral memory (as established through previous learning, or perhaps obtained by cultural transmission). The agent should obviously make use of foreseeable benefits to fitness, automatically. No $f_{est}$-driven stochastic variation is needed for that. The $G$ loop of Figure 1 should handle only those parts of structural change that have unknown effects.

Although any internally generated behavior by an agent could be seen as agency in the wide sense, the specific behavior produced through the $G$ loop can be seen as agency in the narrow sense. Agency in the narrow sense involves an internal, intrinsic overall goal of the agent (the goal of obtaining high $f_{est}$) in addition to a rather special stochastic process for generating behavior. This process generates new

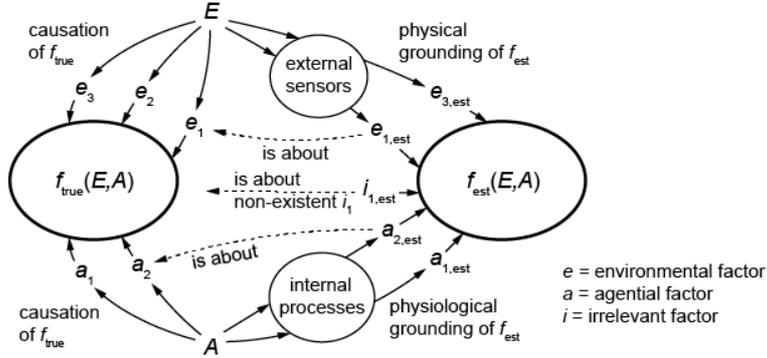

Figure 2: Expansion of the relationship between $f_{est}$ and $f_{true}$ into factors $e$ deriving from the environment $E$, factors $a$ deriving from the agent $A$, and irrelevant factors $i$ (for $f_{est}$ only)

internal structure (and the behavior that derives from that) in an unpredictable way (because of the stochasticity) without being completely random (because the stochasticity is continually driven by the overall goal of high $f_{est}$). Elsewhere I argue that this agency in the narrow sense is the source of the intrinsic goal-directedness and behavioral freedom that living organisms appear to have (van Hateren, 2013).

For the topic of this article, the most interesting part of the theory in Figure 1 is the establishment of reference and aboutness, at least in primordial form. The internal $f_{est}$ is somewhat arbitrary, because the factors that are taken into account for making a reasonable estimate of the value of $f_{true}$ are not fixed, and indeed are likely to vary between members of the same species. Moreover, $f_{est}$ itself is part of the agent, which implies that also $f_{est}$ is subject to structural change through evolution and learning. Nevertheless, $f_{est}$ still must refer to $f_{true}$, be about $f_{true}$, because agents that completely lack this connection are very unlikely to exist. They would not have survived previous natural selection, because they would have been outcompeted by agents with adequate $f_{est}$. Even weak (inaccurate) aboutness has low probability, because such agents would at best belong to a small minority heading for extinction. As I will argue in the next two sections, the fact that $f_{est}$ stands for, refers to, and is about $f_{true}$ is the source of all aboutness of specific variables and symbols in the agent.

## 2.2 Details of Aboutness

Both $f_{true}$ and $f_{est}$ have a complex structure, as illustrated in Figure 2. As mentioned above, $f_{true}$ arises in nature through the interaction of environmental factors ($e_j$, arising from the environment $E$) and agential factors ($a_j$, arising from the agent $A$). Together these factors, and their history up to the present (such as retained indirectly in the genetic or learned memory of agents), physically cause $f_{true}$ in a rather complex way. Also $f_{est}$ is likely to have a complex structure. Although a complete model of $f_{true}$ is infeasible for the agent, there is an evolutionary drive to make the modeled $f_{est}$ as good as possible given the resources of the agent. The better the value of $f_{est}$ mimics the value of $f_{true}$, the better the mechanism of the $G$ loop in Figure 1 can work to increase $f_{true}$. The only way by which $f_{est}$ can faithfully mimic the value of $f_{true}$ across a wide range of environmental conditions is by mimicking at least some of the structure of $f_{true}$ in its own structure, even if only very approximate.

An agent has no direct access to the factors that may be relevant for modeling fitness. External factors that derive from $E$ must be sampled by environmental sensors. Accessing internal states and processes may also require sensors, or at least specialized subprocesses making implicit estimates of the relevant factors. Together, the estimated environmental factors $e_{j,est}$ and agential factors $a_{j,est}$ are the physical and physiological factors producing (causing) $f_{est}$, through an internal process in the agent. However, not all of these factors need to be relevant for $f_{true}$, that is, there may be errors in the structure of $f_{est}$. Such factors ($i_{j,est}$ in the figure) may derive from irrelevant parts of $E$ or $A$, or perhaps may be primarily random and self-generated. Although there is evolutionary pressure to make $f_{est}$ free from errors, there is no guarantee that $f_{est}$ will be optimal in that sense: evolution only requires "good enough" and not "flawless" or "optimal". Some errors may be inevitable if they are causally coupled to factors that are more important and if there is no simple (feasible) way by which the agent can separate them. Moreover, what is an error today may turn out to be an asset tomorrow, and vice versa.

The aboutness of $f_{est}$ with respect to $f_{true}$ (Figure 1) produces the aboutness of factors $e_{j,est}$ and $a_{j,est}$ with respect to corresponding $e_j$ and $a_j$ (broken arrows in Figure 2). However, this also points to several ways reference can be missing or misdirected. First, there may be relevant factors for $f_{true}$, such as $e_2$ in Figure 2, that are completely missed in the estimated $f_{est}$ – perhaps they were not yet discovered as a factor, or perhaps it would be too complicated and take too many resources to incorporate them. One might call this incomplete or missed reference. Second, an estimate, such as $e_{1,est}$ of $e_1$, may vary in its level of accuracy, or vary in how adequately its functional incorporation in the structure $f_{est}$ reflects its true role in the structure of $f_{true}$. If it is not very accurate, the reference is partial at best. Finally, a factor may be false or irrelevant. In that case, it still refers to $f_{true}$, because it inherits its aboutness from the overall aboutness of $f_{est}$ with respect to $f_{true}$ (as depicted in Figure 1). But the reference is then to a wrong or non-existing factor, such as $i_1$ in Figure 2.

The aboutness as illustrated in Figure 2 explains several puzzling aspects of aboutness. First, aboutness is different from a simple physical connection, and in fact somewhat fuzzy. Although both $f_{true}$ and $f_{est}$ are physically connected to $E$ and $A$, and thereby indirectly to each other, there is some arbitrariness in this connection because of the (relative) arbitrariness of $f_{est}$. Complex estimators like $f_{est}$ are likely to be somewhat flexible, with different variants showing similar performance. This implies that the estimates of specific factors, such as $e_{1,est}$, are also flexible. Aboutness and reference are therefore flexible and fuzzy as well. Another puzzling aspect of aboutness is the fact that it can be in error, pointing to false or even non-existent factors (Christiansen & Chater, 1993; the classical example in mental aboutness is thinking about a non-existing animal like a unicorn). As can be seen, this possibility of error is built into the concept right at the low level of the factors depicted in Figure 2. How low-level factors can be combined into high-level symbols, and how these symbols can be stabilized, is discussed in the section below.

## 2.3 Stabilization of Symbols

A primary problem for an agent is that a literal model of $f_{true}$ is infeasible for any realistic situation, when a complex agent has to deal with a natural environment filled with other complex agents. Constructing an $f_{est}$ that is sufficiently adequate for the $G$ loop of Figure 1 therefore requires heuristic strategies, using short-cuts and rules of thumb. One strategy that is likely to be useful is combining elementary factors of $E$ and $A$ into more abstract compounds that can be tested for actual utility. For example, several factors (or compounded factors) may be combined into a compounded factor signifying "danger". Such a compounded factor may be useful when it approximately corresponds to a conjectured compounded

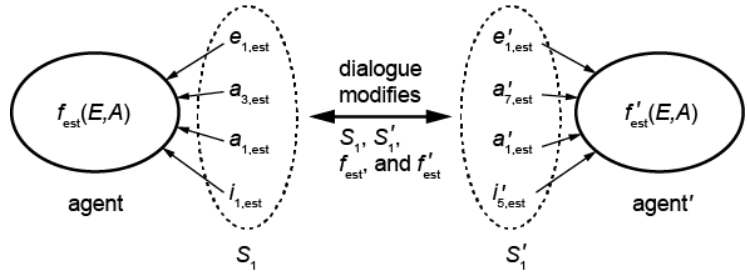

Figure 3: Elementary factors constituting $f_{\text{est}}$ can be compounded into symbols $S$, which can be synchronized and aligned by communication between agents

factor of $f_{\text{true}}$ that explains why $f_{\text{true}}$ decreases when "danger" increases – recall that $f_{\text{true}}$ is the expected rate of reproduction, which indeed temporarily decreases during danger, even when the danger, in retrospect, did not actually result in damage and the realized rate of reproduction was not compromised. Heuristic strategies may thus utilize rather abstract and somewhat fuzzy factors. Conventionally, such factors are called symbols. Symbols are about corresponding aspects of $f_{\text{true}}$, and thereby indirectly about the reality of $E$ and $A$ as causally connected to $f_{\text{true}}$. For example, the agent on the left in Figure 3 has formed a symbol $S_1$ (signifying "danger" for that agent) from several simpler factors. Such factors may include irrelevant ones, that is, factors that the agent believes to signify danger without them actually doing so.

Symbols formed by a single agent are likely to be somewhat arbitrary, because there are many ways to identify and combine factors contributing to a symbol like danger. There are also many ways to include irrelevant or marginally relevant factors. Single agents have only limited opportunities and resources to verify the accuracy and effectiveness of such symbols. In species that are sufficiently social and cooperative, there may be an evolutionary advantage to communicate symbols, for example a symbol signifying impending danger. However, communicating symbols is only useful if the symbols of different agents are approximately aligned. Therefore, communication will drive synchronization and stabilization of symbols (Steels, 2008). For example, the agent on the right in Figure 3 has a symbol $S_1'$ that signifies for that agent "danger" in a roughly similar, but not identical way as $S_1$ for the left agent. If communicating "danger" produces, on average, joint fitness benefits for both agents, then there is an incentive to synchronize the composition of the symbols, by reducing the differences and perhaps getting rid of clearly irrelevant factors. Moreover, the utility of the symbol thus defined for modeling the typical $f_{\text{true}}$ of the agents' species is more easily checked if many agents are sharing the symbol. This implies that there is also a drive to modify symbols such that it increases their utility for accurately estimating $f_{\text{true}}$.

Symbols can become considerably more useful if they are incorporated into a symbolic system such as human language. Such systems not only facilitate synchronizing and modifying symbols, but also enable the additional efficacy and flexibility of symbolic expressions. Moreover, an environment that includes symbol-using agents (partly responsible for $f_{\text{true}}$) implies that models of symbol-using agents should be incorporated into $f_{\text{est}}$ (van Hateren, 2014). It thereby enables symbols (as part of $f_{\text{est}}$) to be about other symbols (as part of $f_{\text{true}}$). These topics obviously form a very large and complex field that is beyond the scope of this article. Nevertheless, the current theory implies that the human grounding of symbols in the form of the meaning of words (and semantics in general) requires more than mere embodiment (or multimodal grounding as, e.g., in Bruni et al., 2014).

## 3. Discussion and Conclusion

Abstracting from the biological context, the primary functional components of the theory presented above are the following. First, there is the strong nonlinearity of exponential growth (or decline) in agent numbers, lying at the heart of reproduction (self-multiplication in the $R$ loop of Figure 1). This nonlinearity depends on a parameter (the reproduction rate $f_{true}$) that is a complex function of a time-varying environment (including other agents) and a gradually changing agent structure. Runaway growth is ultimately stopped by an automatic reduction of $f_{true}$ when resources become limiting. On average, the mechanism of the $R$ loop results in agents that are sufficiently well matched to the environment to reproduce at least as fast as they die. The process therefore appears to optimize $f_{true}$, but only when viewed retrospectively from an external point of view. There are no actual, explicit goals built into the process or into the agents themselves. The process naturally arises from the capability to reproduce with small changes, combined with the inevitable scarcity of resources.

The strong nonlinearity driving the process enables an extension that utilizes stochastic change in the agent's structure that is driven (modulated) by an agent-made estimate of $f_{true}$. This estimate, $f_{est}$, also depends on environment and agent structure, but in a way that is fully defined and controlled within the agent itself. Because of the nonlinearity associated with reproduction, agents with an $f_{est}$-driven loop typically obtain higher $f_{true}$ than agents without an $f_{est}$-driven loop obtain. The extension is therefore evolvable and sustainable by the basic $f_{true}$-optimizing process. The theory requires that $f_{est}$ has evolved to be a reasonable estimate of $f_{true}$, that is, the values of $f_{est}$ and $f_{true}$ are likely to be similar to one another. Although $f_{true}$ and $f_{est}$ are thus coupled, this coupling is not hard-wired, because the way $f_{true}$ is produced (primarily through external processes) is mostly independent of the way $f_{est}$ is produced (through processes evolved and learned within the agent). When the agent induces structural changes within itself that increase $f_{true}$, that is likely (though not guaranteed) to increase $f_{est}$ as well. Similarly, when the agent induces structural changes within itself that increase $f_{est}$, that is likely (though not guaranteed) to increase $f_{true}$. The strength (likelihood) of this co-varying of $f_{est}$ and $f_{true}$ is itself a feature that has been optimized by previous evolution and is maintained by continued evolution.

The coupling between $f_{est}$ and $f_{true}$ implies that agents must have an intrinsic drive to increase $f_{est}$, because that will increase $f_{true}$, on average, and increasing $f_{true}$ is the main feature of the basic optimization process. Because $f_{est}$ is an internal process within the agent, increasing $f_{est}$ is an actual goal of the agent (van Hateren, 2015). It is in fact the agent's ultimate goal, to which all more specific sub-goals (tasks) of an agent are subordinate. Sub-goals are always serving $f_{est}$, but they are not guaranteed to serve $f_{true}$, because the coupling between $f_{est}$ and $f_{true}$ is not perfect. For the topic of this article, the most important consequence of this theory is that $f_{est}$ is about $f_{true}$. The aboutness arises from the fact that $f_{est}$ and $f_{true}$ are tightly coupled, but in a somewhat arbitrary way that is under control of the agent through the way it constructs $f_{est}$.

The primary aboutness produced by the internalized version of fitness within the agent (Figure 1) is inherited by the factors of agent and environment (including other agents) that compose $f_{est}$ (Figure 2). Such factors may or may not correspond accurately to factors composing $f_{true}$, leading to varying degrees of accuracy of aboutness and the possibility of error (such as reference to non-existent items). Straight-forward extensions of the basic theory consist of combining factors into more abstract symbols and of synchronizing and stabilizing symbols amongst communicating agents (Figure 3).

As will be clear from the above summary, the latter parts of the theory – defining and stabilizing symbols through physical and social grounding (Vogt, 2002; Cangelosi, 2006; Coradeschi et al., 2013) and organizing them in a symbolic system – may be complex, but do not require anything that goes beyond existing robotic technology. Indeed, Steels (2008) has shown that letting robots communicate about colored samples can produce a consistent and stable framework for the meaning of the symbols used. The technologically more challenging part is the foundation of the theory, which requires a system with inherent exponential growth that can continually utilize an estimate of its own fitness, $f_{est}$, for modulating stochastic change of its structure.

In evolutionary robotics (Floreano, 2010; Bongard, 2013), it is normally the experimenter who decides how $f_{true}$ is defined. For example, robotic designs that can move faster across a plane (Lipson & Pollack, 2000) are granted more reproduction, and therefore the possibility to mutate into forms that might move even faster. In the theory presented above, such a goal would only be a sub-goal, serving the ultimate goal of fast reproduction (high $f_{true}$ and $f_{est}$). Evolving and learning biological agents can therefore switch sub-goals whenever such switching is technically possible (given physical and physiological constraints and constraints on the pathways open for evolution and learning) and whenever it increases their rate of reproduction. They can therefore autonomously abandon a sub-goal like moving fast and switch to new and unpredictable sub-goals when their reproduction is served in that way.

A full implementation following the organic version of the theory would require solving several challenging problems. First, the problem of how to realize fully autonomous reproduction, including autonomous gathering of the materials and energy needed for reproduction. Second, the problem of designing an initial structure that has a sufficiently open-ended evolvability such that it can serve the basic optimization mechanism of the $R$ loop of Figure 1, and can continue to do so indefinitely. Finally, the problem of constructing the initial version of an evolvable, learning, and $f_{true}$-mimicking $f_{est}$ that can stochastically modulate structural change and thereby serve the extended optimization mechanism of the $G$ loop of Figure 1. However, successfully solving these problems implies that the evolving artificial agents will occasionally give rise to exponential growth in numbers – a plague. Waves of extinction will also occur occasionally. This would not only imply a waste of resources, but also competition with organic life, including humans. The artificial agents' overall goal of fast reproduction is independent of and therefore likely to be unaligned with human goals. The same applies to the sub-goals composing the overall goal. Thus even if autonomous reproduction (and a functioning $f_{est}$) would be feasible – either with materials similar to or different from current biological ones – there would be compelling reasons not to implement the theory because there would be no alignment (that is, likely conflict) with human purposes.

Can the theory also work with agential goals that are kept constrained to goals that serve human interests? Clearly, the basic mechanism (loop $R$ in Figure 1) could work like that, as the example of quickly moving robots given above illustrates. However, agents evolving through such a mechanism do not have the symbol grounding as proposed here, because they lack the $G$ loop and $f_{est}$. Their goals and symbols therefore remain parasitic on human goals and symbols, and the symbols lack autonomous aboutness. The $G$ loop depends on a subtle stochastic process, because the modulated stochasticity does not change fitness directly, but only through a mechanism depending on second-order statistics (variance of structural change). Therefore, it can only function if it is driven by a strong nonlinearity. In organic life, this nonlinearity comes from the exponential growth in numbers implied by self-reproduction. This produces not only strong selection (nearly all variants die, only very few can continue), but also prevents complete extinction by quickly multiplying the numbers of the few successful variants.

If the autonomous goal of self-reproduction (high $f_{est}$) is unacceptable, as suggested above, then $f_{est}$ must be replaced by another goal, say $g_{est}$. In principle, this is possible (for an example of a simulation with an arbitrary goal see figure 1b in van Hateren, 2015). For example, $g_{est}$ might by defined as high when a robot can move quickly across a place. Then low $g_{est}$ (slow moving) would imply an increased mutation rate ($\sim 1/g_{est}$ as in Figure 1). However, when $g_{est}$ is high, that is, when the task is performed well, such successful variants must still be selected and be protected from extinction by letting them multiply. In other words, $g_{est}$ must be coupled not to a $g_{true}$, but to a genuine reproductive fitness, $f_{true}$ – either that, or multiplying must be controlled purely through human intervention[1]. But for autonomous agents, $g_{est}$ and $f_{true}$ will generally have a rather different value and structure if the assigned goal $g_{est}$ has little to do with reproduction. The aboutness of Figures 1 and 2 will then be broken.

A second issue is that $g_{est}$ is as much a part of the agent's structure as anything else. In an evolving or learning agent, it is therefore likely to change along with the rest of the agent, unless it could somehow be isolated. In organic life, $f_{est}$ is presumably distributed throughout the agent (see Section 2.1), and isolating it from change is then not an option. It is not clear if sustained isolation could be realized in artificial agents, but it is at least conceivable. However, isolating $f_{est}$ from change may hinder the synchronization of symbols as depicted in Figure 3, and thereby prevent the establishment and maintenance of a useful symbolic system. Moreover, if such an isolation turns out to be infeasible in practice, then the goal $g_{est}$ will eventually align itself with $f_{true}$, because the latter drives the basic Darwinian optimization process. Then $g_{est}$ will change into $f_{est}$, gradually, but inevitably. Such change may happen quite slowly if the agent has little or limited intelligence, such as in a bacterium or insect. But it will happen much faster when agents are more intelligent, because high intelligence requires the capacity to change sub-goals continually and to expand the space of possible sub-goals into unpredictable directions – not just on the timescale of evolution but also on the timescale of the individual's lifetime.

The conclusion from the above considerations is that it is not clear how goals can be kept constrained to serve human goals (for similar concerns see Bostrom, 2012). Even if it were possible, for example by isolating $g_{est}$ from the agent's structures involved in learning, it is not clear how symbol grounding in the form of aboutness could then be obtained, because $g_{est}$ would be unrelated to $f_{true}$. The primary question is, then, if another solution than the organic one can be found that retains the special stochastic properties of the $G$ loop and that would produce aboutness without the issues sketched here. It will be interesting to see if this problem has a solution, and if human ingenuity can find it.

---

[1] Self-reproduction and alignment with human goals could be controlled and enforced, as the case of animal husbandry shows. However, if the agents involved would have a level of intelligence that requires symbolic reasoning, enforcement would presumably be unethical as well as resisted by the agents, and goals might be actively sabotaged.